\documentclass[10pt,twocolumn,letterpaper]{article}

\usepackage{iccv}
\usepackage{times}
\usepackage{epsfig}
\usepackage{graphicx}
\usepackage{amsmath}
\usepackage{amssymb}
\usepackage{bbm}
\usepackage[ruled,vlined]{algorithm2e}
\usepackage{booktabs}
\usepackage{subfigure}

\usepackage[breaklinks=true,bookmarks=false]{hyperref}

\iccvfinalcopy 

\def\iccvPaperID{7325} 

\ificcvfinal\fi

\begin{document}

\title{Weakly Supervised Keypoint Discovery}

\author{Serim Ryou\\
California Institute of Technology\\
{\tt\small sryou@caltech.edu}
\and
Pietro Perona\\
California Institute of Technology\\
{\tt\small perona@caltech.edu}
}

\maketitle
\ificcvfinal\thispagestyle{empty}\fi

\begin{abstract}
   In this paper, we propose a method for keypoint discovery from a 2D image using image-level supervision. Recent works on unsupervised keypoint discovery reliably discover keypoints of aligned instances. However, when the target instances have high viewpoint or appearance variation, the discovered keypoints do not match the semantic correspondences over different images. Our work aims to discover keypoints even when the target instances have high viewpoint and appearance variation by using image-level supervision. Motivated by the weakly-supervised learning approach, our method exploits image-level supervision to identify discriminative parts and infer the viewpoint of the target instance. To discover diverse parts, we adopt a conditional image generation approach using a pair of images with structural deformation. Finally, we enforce a viewpoint-based equivariance constraint using the keypoints from the image-level supervision to resolve the spatial correlation problem that consistently appears in the images taken from various viewpoints. Our approach achieves state-of-the-art performance for the task of keypoint estimation on the limited supervision scenarios. Furthermore, the discovered keypoints are directly applicable to downstream tasks without requiring any keypoint labels.
   
\end{abstract}

\section{Introduction}

\begin{figure}[t]
\centering
\includegraphics[width=8cm]{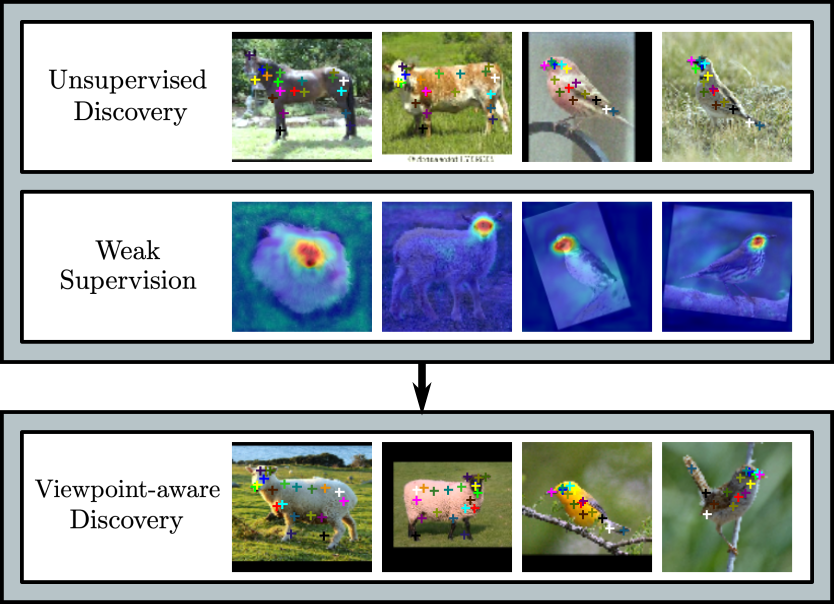}
\caption{Overview: Our approach uses the unsupervised method to discover diverse keypoints and image-level supervision to localize the discriminative parts. By using the keypoints learned from weak-supervision to infer the viewpoint of a target instance, our model can successfully discover semantically consistent parts for instances facing in different directions. }
\label{fig:summary}
\end{figure}

Keypoints are a convenient intermediate representation towards final tasks, such as action recognition~\cite{Li_2020_CVPR}, fine-grained classification~\cite{BransonVBP14,GuoWACV19}, face identification~\cite{xie_eccv18}, and person re-identification~\cite{Su_2017_ICCV,zhaoCVPR17}. However, collecting keypoint annotations is labor-intensive and time-consuming compared to the image-level or bounding box annotations. Recently, unsupervised keypoint discovery~\cite{JakabNeurips18, jakab20self-supervised,Lorenz19,ZhangKptDisc18} has been proposed to reduce the annotation effort and has shown successful results for the images with aligned instances and human pose. However, these methods struggle to find consistent keypoints when the target objects have severe viewpoint and shape variation. On the other hand, weakly-supervised learning methods on object localization~\cite{jie2017deep,cvpr2016_zhou,Oquab_CVPR15,SinghL17,Tang_2019,Wang_ECCV14} easily identify the discriminative parts of the target object by using the features trained from the deep neural networks with class labels. In this work, we propose a weakly-supervised keypoint discovery method by exploiting the image-level supervision to guide the network to discover discriminative parts and a viewpoint. To diversify keypoints, our method adopts the unsupervised methods~\cite{JakabNeurips18,jakab20self-supervised,Lorenz19,ZhangKptDisc18} which learn the image generation conditioning on the structural bottleneck. Figure~\ref{fig:summary} illustrates the overview of our method. 


Most of the current unsupervised keypoint discovery approaches~\cite{JakabNeurips18,jakab20self-supervised,Lorenz19,ZhangKptDisc18} share the idea of disentangling the representation of appearance and structure. Given a pair of source and target images with structural transformation, these works~\cite{JakabNeurips18,jakab20self-supervised} extract a keypoint information from a target image and the appearance representation from a source image. With the appearance feature from the source image, the network is trained to reconstruct the target image by using the keypoint bottleneck computed for its structural representation. These methods automatically discover semantically meaningful parts for the images with aligned instances. However, empirical results show that when the instance has high viewpoint variation, the model fails to find semantically consistent parts. Specifically, animals usually have diverse poses with high appearance and viewpoint variation. The discovered keypoints from animal images show a high correlation on the spatial coordinates and lose the semantic correspondence across different images.

On the other hand, weakly-supervised learning methods on object localization~\cite{jie2017deep,cvpr2016_zhou,Oquab_CVPR15,SinghL17,Tang_2019,Wang_ECCV14} easily identify the most discriminative parts, while suffering from localizing only the dominant region (\eg face of animal). Our method exploits this idea to discover the discriminative parts when the target instance has a large viewpoint and shape variation. We use the part-based representation by extracting the local features from the discovered keypoint locations and train these features to predict the image-level labels. The parts discovered from this process are simultaneously used for the unsupervised image reconstruction task as well. 

After adopting these two approaches, we observe that the discovered keypoints still show a high spatial correlation between different parts that have a similar appearance, \eg parts from the torso or front and back legs. To resolve this issue, we propose a viewpoint-based equivariance constraint, where the keypoint representation should move according to the structural deformation. Unlike using the equivariance constraint on all pairs of images~\cite{Lorenz19,ZhangKptDisc18}, our method applies this constraint only to the viewpoint-augmented images. We use discovered parts from the image-level supervision to infer the viewpoint of an instance and enforce the equivariance constraint based on the model prediction.

We evaluate our method in various experimental settings. To compare with existing keypoint discovery methods, we first test on datasets with small viewpoint variation, \eg facial keypoint, and animals with a consistent viewpoint. Moreover, we demonstrate the robustness of our method to a large viewpoint and appearance variation by applying it to challenging datasets that include diverse species of animals. When trained with datasets with large shape diversity, our model can handle high appearance variation and discover the keypoints from the images with unseen categories. For both cases, our method achieves state-of-the-art performance in the limited supervision scenarios. Finally, we analyze the distribution of the discovered keypoints and demonstrate its representation power by applying it to a simple behavior classification task.

\begin{figure*}[t]
\centering
\includegraphics[width=16cm]{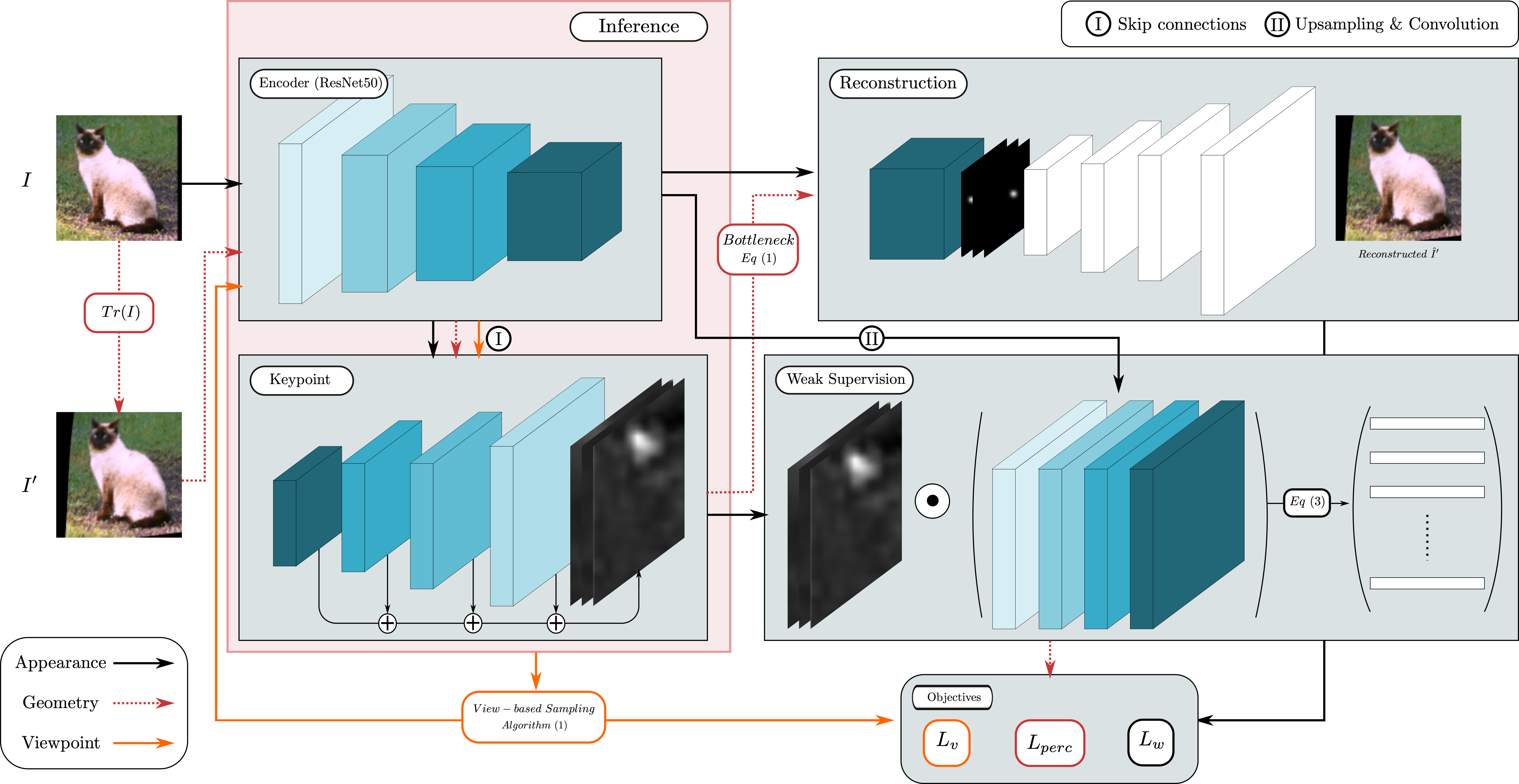}
\caption{System Outline: Our model uses the shared encoder for various tasks: image reconstruction, keypoint estimation, and classification. The bluish same block color represents the features generated from the shared weights of ResNet-50. Black, red, and orange arrows represent the appearance, geometry, and viewpoint stream, respectively. }
\label{fig:pipeline}
\end{figure*}

\section{Related Work}


Our goal is to build a keypoint discovery model which is robust across viewpoints by incorporating information from pose estimation, weakly-supervised learning, and unsupervised keypoint discovery. \\


\noindent\textbf{Keypoint Estimation}
Keypoint estimation is a problem of localizing a predefined set of keypoints from an input image. Pose is a convenient intermediate representation for various applications. Applications range from human pose estimation~\cite{CPN17,Newell2016StackedHN,Ryou_2019_ICCV,Tang_2018_ECCV,PoseMachines16}, facial landmark detection~\cite{Belhumeur11,RCPR13,CaoFace14,ZhangECCV14} to animal pose estimation~\cite{BransonVBP14,GuoWACV19,Mathisetal2018}. With the development of fully convolutional neural networks~\cite{FCN_segmentation17}, the pose estimation community gained huge success by estimating the part locations using a heatmap, where each location is encoded as a 2D gaussian map centered at each body part location. Specifically, most of the existing approaches exploited iterative refining steps~\cite{Newell2016StackedHN,PoseMachines16}, multi-scale information~\cite{CPN17}, and learning signals~\cite{CPN17,Ryou_2019_ICCV} for further improvement. \\

\noindent\textbf{Unsupervised Keypoint Discovery}
Despite the success in estimating the part location with supervision, one of the major drawbacks is that it requires a huge amount of annotations. 
Recently, methods to discover the landmarks without manual annotation emerged with the shared idea of image deformation and reconstruction~\cite{JakabNeurips18,Lorenz19,ZhangKptDisc18}. 
Jakab~\etal~\cite{JakabNeurips18} used the discovered keypoints as a geometry bottleneck for conditional image generation.
Zhang~\etal~\cite{ZhangKptDisc18} proposed an autoencoder-based architecture by explicitly using the feature representation from the discovered keypoints. Lorenz~\etal~\cite{Lorenz19} disentangled appearance and structure representation by exploiting the shape and appearance transform separately. Jakab~\etal~\cite{jakab20self-supervised} incorporated prior knowledge about the pose of the target by using unpaired keypoint data from existing datasets to discover the keypoints for other datasets within the same domain. Our work does not require any prior knowledge about the structure of the target instance and tackles a more challenging problem where the target instance has a large viewpoint and shape variation. \\



\noindent\textbf{Weakly-supervised Learning}
Weakly-supervised learning methods have been adopted for various vision tasks including object localization~\cite{jie2017deep,Oquab_CVPR15,SinghL17,Tang_2019,Wang_ECCV14}, semantic segmentation~\cite{Huang_CVPR18,pathakICCV15ccnn}, and semantic matching~\cite{NovotnyLV17}. Zhou~\etal~\cite{cvpr2016_zhou} proposed Class Activation Map (CAM) for object localization only with the class labels and demonstrated that the image-level supervision gives a cue to find the most discriminative region of the objects. At the same time, weakly-supervised methods struggle from predicting only the dominant parts rather than the entire object. While previous work aims to resolve this issue by manipulating the image patches~\cite{SinghL17} or iteratively refining the classifiers~\cite{jie2017deep,Tang_2019}, our work exploits this idea to discover the consistent discriminative parts.  \\






\noindent\textbf{Part-based Representation} 
Part-based features have been useful representations for many computer vision applications, especially for the tasks of disambiguating the marginal visual differences: fine-grained classification~\cite{BransonVBP14,GuoWACV19,Sun_ECCV18,Zhang_FG_2019} and facial identification~\cite{xie_eccv18}. Fine-grained image classification works use keypoint information either by explicitly estimating the keypoint locations using the ground truth~\cite{BransonVBP14,GuoWACV19} or implicitly discovering the parts~\cite{Sun_ECCV18,Zhang_FG_2019}. While the latter works also automatically learn the parts, the keypoints are the byproduct of the final task, thus they do not measure the semantic consistency over different images. On the other hand, the goal of our work is to discover the keypoints which are consistent over different images and species. 

Recent works on action recognition~\cite{Du_2015_CVPR,Li_2020_CVPR,conf/cvpr/WangWY13} use only the coordinate-based representation as an input to the activity classification. Our work shows potential for the discovered keypoints to be used as an input representation to the simple behavior classification tasks. 




\section{Method}


The goal of our work is to discover the keypoints for the instances having various viewpoints. Existing methods on object localization with image-level supervision~\cite{SinghL17,cvpr2016_zhou} easily identify the most discriminative parts. Recent works on unsupervised keypoint discovery have been successful on images with aligned instances like facial keypoints, and have shown potential on animals with a consistent viewpoint~\cite{Lorenz19,ZhangKptDisc18}. 
Our work is based on these two approaches by detecting the coarse parts with image-level supervision and discovering the fine parts with an unsupervised reconstruction module. The overall pipeline of our method is shown in Figure~\ref{fig:pipeline}. In this section, we explain the architecture of each module. 

\subsection{Unsupervised Keypoint Discovery}
In our experimental setting, we use ResNet-50~\cite{He2016DeepRL} as a backbone for keypoint discovery, image reconstruction, and the weak-supervision modules. The convolution feature blocks $\{C_1, C_2, C_3, C_4\}$ from each resolution of ResNet-50 are used throughout all modules. 

Given a pair of source and target images $(I, I')$ with structural transformation, the model learns to reproduce the target image by using the appearance representation from the source image and the geometry information from the target image. The dotted red line in Fig~\ref{fig:pipeline} represents the pipeline for the unsupervised keypoint discovery.

Specifically, the source image $I$ and the transformed target image $I'$ are fed to the shared image encoder. Here, we use Thin Spline Transformation (TPS)~\cite{Spline_Jean77,Wahba90a} as a transformation function to generate a target image. The encoder generates the feature representations $\{C_1, C_2, C_3, C_4\}$ and $\{C'_1, C'_2, C'_3, C'_4\}$ for the source and target images, respectively. The feature $C_4$ from the source image is used as an appearance feature and the features generated from the target image are used to extract the geometry information from the keypoint module.

\noindent\textbf{Keypoint module} 
We use the GlobalNet architecture from Cascaded Pyramid Network (CPN)~\cite{CPN17}, which exploits multi-scale features from ResNet-50~\cite{He2016DeepRL}, for the keypoint discovery module. From each resolution of convolutional blocks $\{C'_1, C'_2, C'_3, C'_4\}$ from the target image $I'$, the network generates heatmaps and upsamples them to the final output size. The sum of the heatmaps from each resolution becomes the final heatmap, which is used for generating a keypoint bottleneck.

\noindent\textbf{Image reconstruction with discovered landmarks}
The geometry information from the target image should capture the structural differences from the source image. We represent the discovered keypoints as a tight geometry bottleneck by generating a gaussian heatmap. From the prediction of the keypoint module, we apply spatial softmax to each channel and use these normalized heatmaps $H_k$ to compute the weighted sum over $x, y$ coordinates to get the $p_k = (u_k, v_k)$ locations for $k=\{1,\dots,K\}$ keypoints. To explicitly localize the target part, we generate 2D Gaussian heatmaps centered on the keypoint locations and these heatmaps become a structure bottleneck $B_k$ for the target instance. 

\begin{align}
    B_k(\textbf{x}) = &\frac{1}{\sqrt{2\pi\sigma^2}} \exp \left( {\frac{- \left\Vert\textbf{x}-p_k\right\Vert^2 }{2\sigma^2}} \right)
\end{align}

The concatenation of the last feature from the encoded source image $C_4$ and the structure bottleneck $B_k$ from the target image becomes an input to the reconstruction module, which consists of convolution and upsampling layers. We feed geometry information to each resolution of the reconstruction module. The architecture details about this module are in the supplementary material. For the image reconstruction, we use perceptual loss $L_{perc}$~\cite{Johnson2016Perceptual}, which compares the features computed from VGG~\cite{VGG14} network $f$ with the target $I'$ and the reconstructed images $\hat{I}'$. 

\begin{align}
    L_{perc} = \sum_i \left\Vert f(I') - f(\hat{I}';B(I'))) \right\Vert_2
\end{align}








\begin{figure}
\centering
\includegraphics[width=8cm]{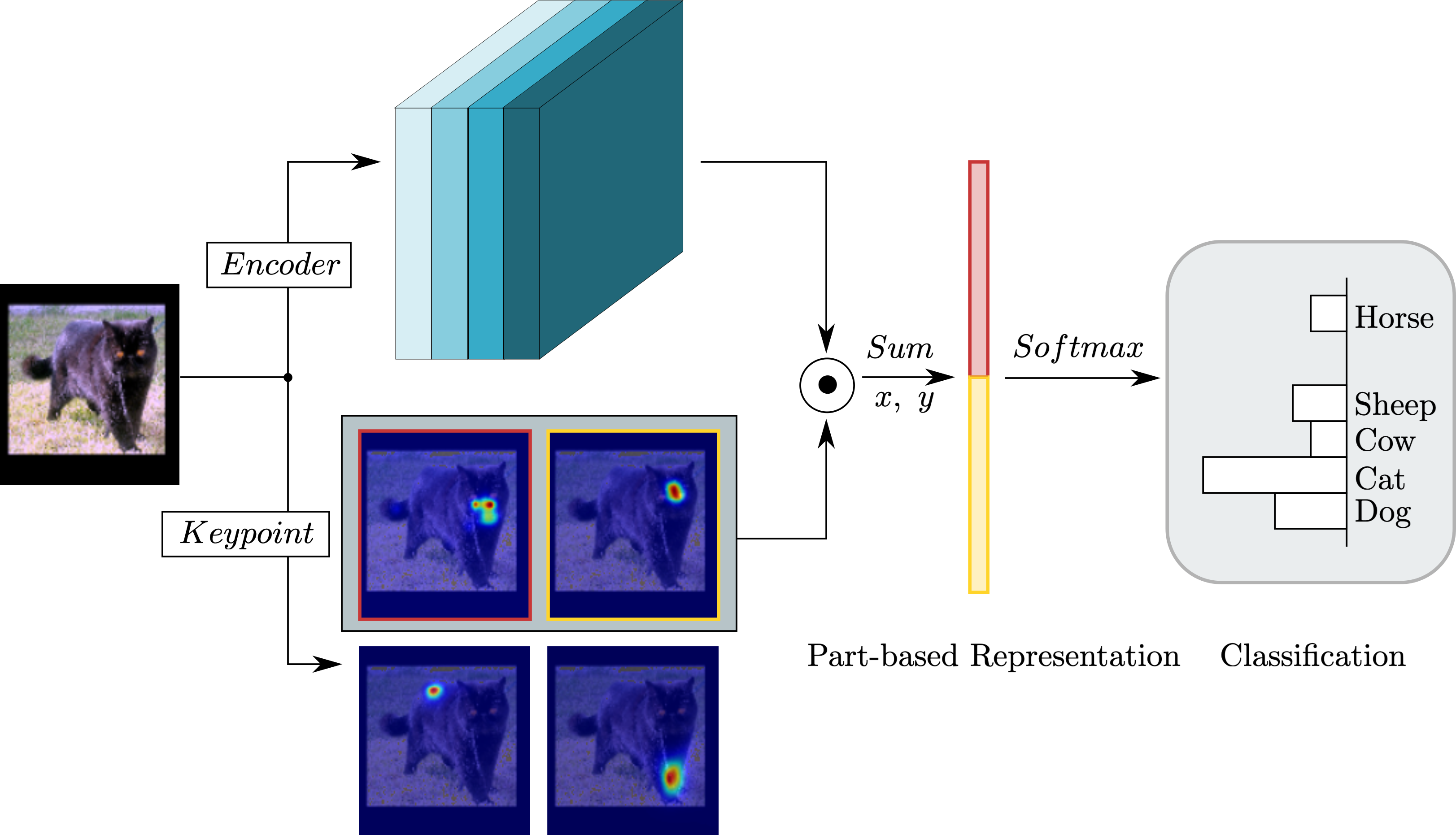}
\caption{Weak-supervision module: We use first $K_w$ keypoints to extract part-based representation from the base features. Concatenated part-based features are fed to predict the image-level supervision task.}
\label{fig:weak_supervision}
\end{figure}

\subsection{Weak Supervision}

We extract the part-based representation from the features obtained by the shared encoder. Since the network loses spatial information as it goes deeper, we upsample features from each convolutional block $\{C_1, C_2, C_3, C_4\}$ to the final heatmap size and apply few convolution layers to reduce the feature dimension. These feature blocks $C$ become the base representation for the image-level supervision tasks. Then, we generate a keypoint-based representation by applying the Hadamard product to the base features and the discovered keypoint heatmap. Here we use the $K_w < K$ number of heatmaps to extract features for the target task. Figure~\ref{fig:weak_supervision} illustrates this process. 

\begin{align}
    h_k = \sum_i \sum_j H_k(i,j) \odot C(i,j)
\end{align}

Each vector represents a localized feature for each keypoint. The concatenated vectors are fed to the final fully connected layer. Here we apply cross-entropy for the classification task as our weak supervision loss. 

\begin{align}
    L_w &= -\sum_i^N y_i \log \hat{y}_i
\end{align}

\subsection{Viewpoint-based Equivariance} 
\begin{figure}
\centering
\includegraphics[width=8cm]{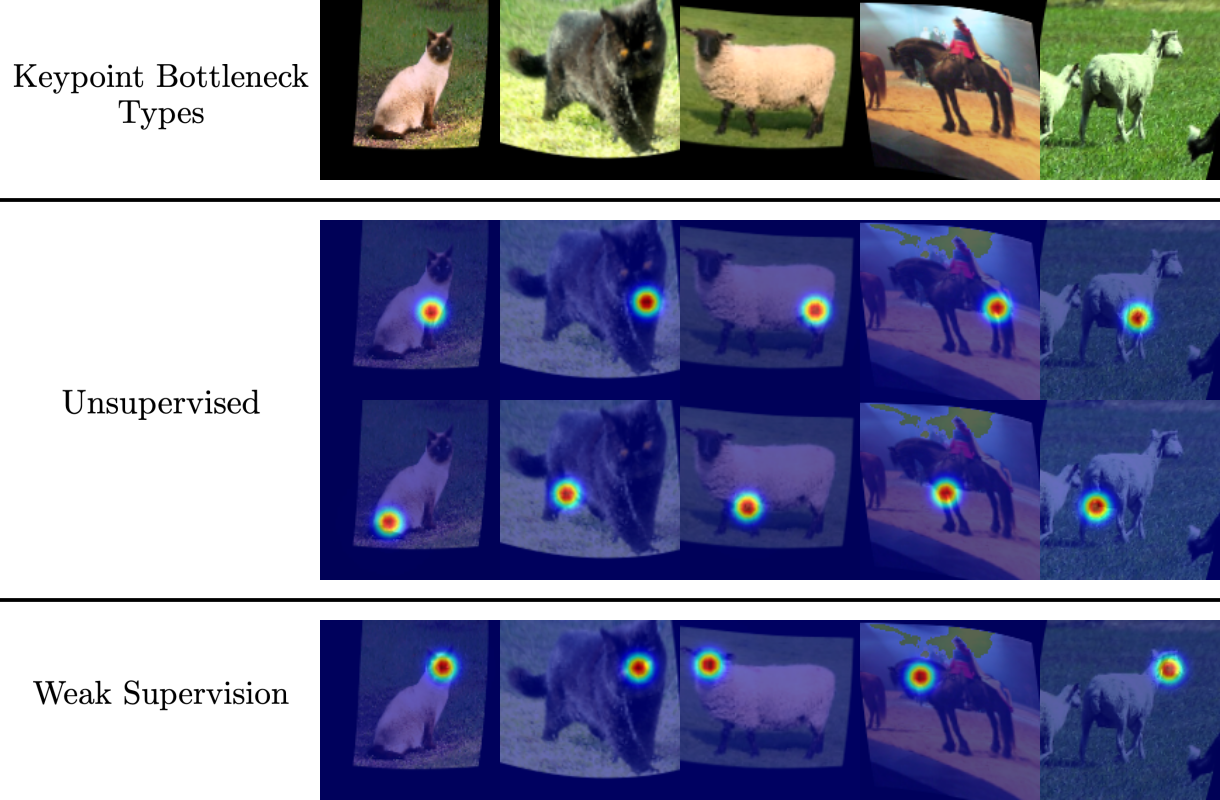}
\caption{Spatial correlation between similar parts. Each row represents the keypoint bottleneck generated from one channel. Keypoints discovered from the unsupervised module predict similar spatial locations for look-alike body parts regardless of the viewpoint of the animal. On the other hand, keypoints discovered from the weak-supervision module consistently find discriminative parts like the face.}
\label{fig:spatial_corr}
\end{figure}

We observe that the discovered keypoints have a spatial correlation on the parts that have a similar appearance, except for the discriminative parts that are tied to the weak supervision module. Figure~\ref{fig:spatial_corr} shows that the keypoints discovered from the torso and legs are predicting similar spatial locations although the animals are facing in opposite directions. In order to resolve this issue, we propose an equivariance constraint based on the model predictions from the same viewpoint to generate the keypoint labels for the opposite viewpoint by data augmentation. Unlike previous works using equivariance constraint on the pair of images~\cite{Lorenz19,ZhangKptDisc18} with small deformation, our work applies it to viewpoint-based data augmentation like flipping. \\

\noindent\textbf{Sampling based on weak supervision}
In order to generate the labels for the viewpoint-augmented image, we have to efficiently sample the images with a shared viewpoint. This sampling process involves the model prediction and the keypoints discovered from the image-level supervision task to infer the facing direction of a target instance. The procedure of sampling and training is explained in Algorithm~\ref{alg:sampling}

\SetArgSty{textnormal}
\begin{algorithm}[t]
    \SetAlgoLined
    \SetKwInput{KwData}{Input}
    \KwData{Images $I$ in a mini-batch at iteration t; Keypoint network $\phi(I)$}
    \KwResult{Images for equivariance loss $I^v$; Corresponding keypoint labels $p^*$}
     \For{$i=1,...,N$} {
      Predict the discovered keypoints $(\textbf{u}_i,\textbf{v}_i) = \phi(I_i)$ \\
      Compute the $x$ variance $s_i = \sigma^2 (\textbf{u}_i)$ \\
      }
      Sample one-side facing images using S \\
      1. Sort S in descending order and choose $N_s$ images with high $x$ variance \\
      2. For $N_s$ images, compute the mean location of $\mu_i = (\bar{\textbf{u}}^k_i, \bar{\textbf{v}}^k_i)$ for $k=\{1,\dots,K_w\}$\\
      3. Pick $N_v$ samples facing the same direction by sorting $\mu$. \\
      4. Generate the view-augmented images $I^v$ by flipping $N_v$ images \\
      5. For $N_v$ images, set equivariance label $p^*$ by flipping the discovered keypoints.  
    \caption{Viewpoint-based Equivariance}
    \label{alg:sampling}
\end{algorithm}

For the samples obtained by this procedure, we apply MSE loss $L_v$ as an equivariance constraint.
\begin{align}
    L_v = \frac{1}{N_v} \sum_i^{N_v} \Vert \phi(I^v_i) - p^*_i \Vert_2
\end{align}







\subsection{Loss}
The final objective function is composed of three parts: the perceptual loss $L_{perc}$ for image reconstruction, image-level supervision loss $L_w$, and viewpoint-based equivariance loss $L_v$. Since the equivariance constraint depends on the model prediction, we adopt curriculum learning~\cite{Bengio2009} for training $L_v$ loss. The hyperparameter settings are in the supplementary material. 

\begin{align}
    L = w_p L_{perc} + w_w L_w + w_v L_v\mathbbm{1}_{\{epoch>n\}} 
\end{align}


\section{Experiments}

In this section, we evaluate our method on keypoint estimation and downstream tasks. First, we show the qualitative results with discovered keypoints on various datasets. 
To quantitatively measure the performance of our method, we evaluate two different experimental settings: linear regression and finetuning. Secondly, we analyze our model output by showing the distribution of the predictions and visualize the pose embedding. We also show the performance on the weak supervision task of fine-grained classification. Finally, we demonstrate the efficacy of our discovered keypoints by directly predicting simple animal behaviors from the discovered keypoints without any keypoint label. 

\subsection{Datasets}
We conduct experiments on various datasets with large viewpoint and appearance variations. To compare with existing unsupervised methods, we run the experiments on the images with a consistent viewpoint (CelebA, CUB) and images with various viewpoints (CUB, AnimalPose, StanfordDogs). In addition, we test our method by applying the discovered keypoints to a simple activity prediction task (DogPart, TigDog). We briefly explain each dataset here. \\

\noindent\textbf{CelebA}
\cite{liu2015faceattributes} is a dataset of 200k facial images with 10k identities. We follow the same training and testing split of~\cite{Lorenz19,ZhangKptDisc18}, which excludes the train and test set from MAFL. We train the linear regressor for 5 keypoints using the MAFL training set 19k images and test on 1k MAFL test set. Since this dataset has a fixed viewpoint with small appearance variation, we only use the image reconstruction loss for training this dataset.


\noindent\textbf{CUB}
\cite{WelinderEtal2010} is a dataset of fine-grained classification of the bird species with 200 categories and 15 keypoint labels. We test on two different settings with the CUB dataset. First, to compare with the unsupervised methods~\cite{Lorenz19}, we exclude the seabird species and align the parity using the visibility of the eye landmark. In addition, we test on the full dataset including the images with all species and various viewpoints by finetuning the keypoint network. 

\begin{figure}
\centering
\includegraphics[width=8cm]{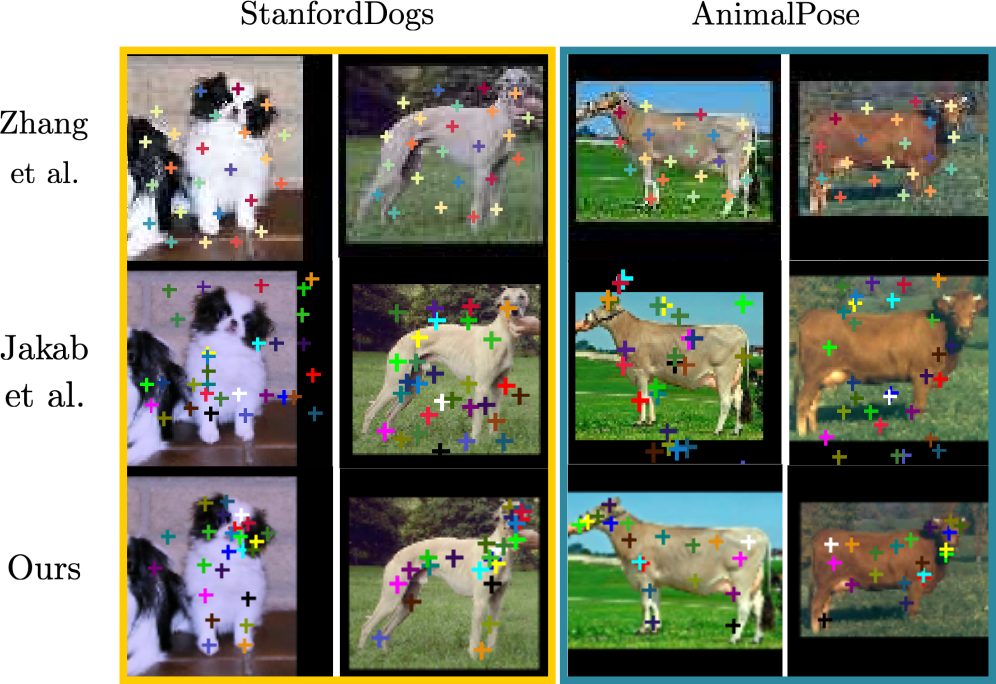}
\caption{Discovered keypoints on the AnimalPose dataset using existing methods~\cite{JakabNeurips18,ZhangKptDisc18}. Unsupervised models either predict similar locations regardless of the semantic parts and viewpoint of an instance or fail to discover semantically consistent parts. }
\label{fig:comparison}\vspace{-0.2cm}
\end{figure}

\noindent\textbf{StanfordDogs}
\cite{Khosla_FGVC2011} is a dataset for fine-grained classification of 120 dog species with 20k images. Recently, the StanfordExtra dataset~\cite{biggs2020wldo} has been released with silhouette and 24 keypoint labels. We evaluate the keypoint performance on the StanfordExtra dataset. 


\noindent\textbf{Animal Pose}
\cite{CaoICCV19} is a dataset for a cross-domain adaptation task with 12 different species of animals. This dataset contains bounding box annotations for 7 animal categories and the pose labels for 5 different species in a total of 6k instances in 4k images. 20 keypoints are labeled for the animals with the pose label.


\noindent\textbf{DogPart}
\cite{Barnard2016} is a dataset for automatic animal behavior classification. This dataset is composed of 10 videos taken from a zoo or indoor environment and each frame is labeled with 3 different posture-based action categories: standing, sitting, and lying. In our experiments, we extract the frames that have keypoint and action labels and loosely crop the bounding box area, which brings to a total number of 1k images.

\noindent\textbf{TigDog}
~\cite{delpero15cvpr} is a dataset for behavior analysis. We use the subset of the action categories that can be identified by each frame: standing, sitting, and rolling for horse images. The process to extract the images is in the supplementary materials and the curated dataset for our experimental setup has around 2k images. \\

\noindent\textbf{Implementation Details}
We set the input image size to 128x128 and the number of keypoints for the weakly supervised task to 5 for all experiments. Our model is based on pretrained model on ImageNet~\cite{imagenet_cvpr09}. For the restricted setting, we do not apply the viewpoint-equivariance loss since there is no viewpoint change in the dataset. We adopt curriculum learning for the full dataset experiments. Hyperparameter settings are in the supplementary material. We discover the same number of keypoints provided by each dataset unless otherwise specified.  

\subsection{Keypoint Estimation}

\begin{figure}
\centering
    \subfigure[CelebA]{\includegraphics[width=8cm]{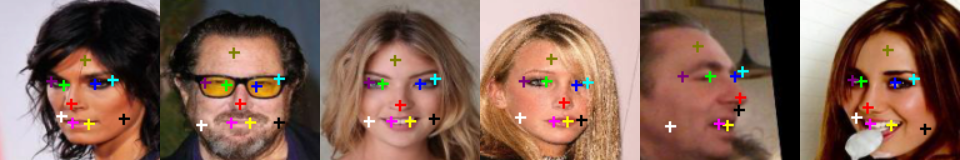}}\vspace{-0.2cm}
    \subfigure[CUB]{\includegraphics[width=8cm]{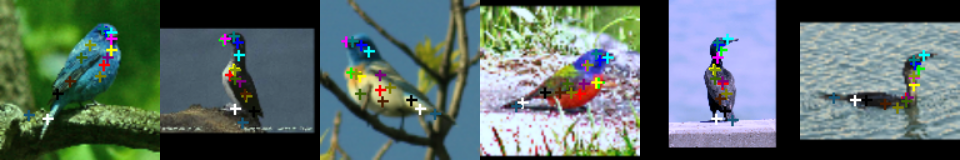}}\vspace{-0.2cm}
    \subfigure[StanfordDogs]{\includegraphics[width=8cm]{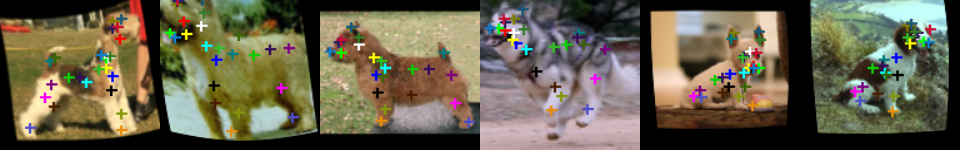}}\vspace{-0.2cm}
    \subfigure[AnimalPose]{\includegraphics[width=8cm]{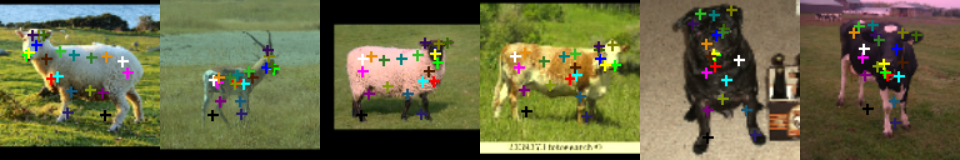}}
    \caption{Qualitative results on various datasets. Our model successfully discovers semantically consistent parts for images with large viewpoint and appearance variations. }
    \label{fig:qualitative}
\end{figure}

\begin{figure*}
\centering
\includegraphics[width=5.5cm]{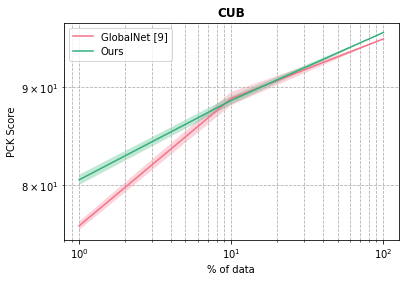}
\includegraphics[width=5.5cm]{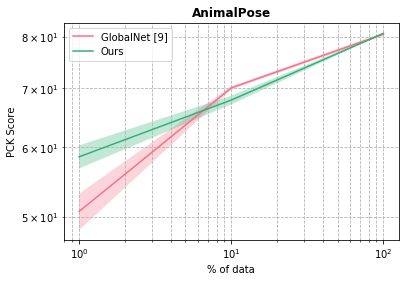}
\includegraphics[width=5.5cm]{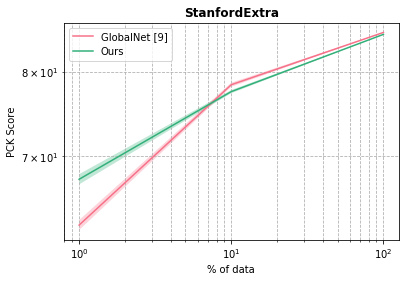}
\caption{PCK score by finetuning the keypoint network with a different amount of supervision on CUB, AnimalPose, and StanfordExtra datasets. Despite the marginal performance differences after using 10\% of supervision, the representation learned from our method gives better performance when there is an extremely limited amount of supervision.}
\label{fig:diff_sup}\vspace{-0.2cm}
\end{figure*}

Figure~\ref{fig:comparison} and~\ref{fig:qualitative} show qualitative results on the discovered keypoints with zero keypoint annotation compared to the result from existing methods (Fig~\ref{fig:comparison}). Our method can successfully discover keypoints when the target instances have large viewpoint variations. To quantitatively measure the performance of our method, we follow the same evaluation protocol from previous unsupervised works~\cite{JakabNeurips18,Lorenz19,ZhangKptDisc18} by learning a linear regressor from the model prediction to keypoint annotations for the viewpoint-constrained datasets. For animal datasets, a simple linear regressor cannot capture the relation between the prediction and the annotations due to significant viewpoint changes across the images. Thus, we finetune the keypoint network and evaluate the keypoint estimation performance by varying the number of keypoint annotations with the supervised models. 


\begin{table}
    \caption{Keypoint estimation performance on the restricted setting. We train a linear regressor from the discovered keypoints for MAFL and CUB datasets. For CUB experiments, we follow the same data extraction step from the paper~\cite{Lorenz19} and show the performance with an edge length of the image as an inter-ocular distance. }
    \label{table:kpt_restricted}
    \begin{center}
    \scalebox{0.8}{%
    \begin{tabular}{lcc}
    \toprule
    Dataset & MAFL & CUB \\
    K & 10 & 10 \\
    \midrule
    Thewlis~\cite{Thewlis_2017_ICCV}  & 6.32 & - \\
    Jakab~\cite{JakabNeurips18} & 3.19 & - \\
    Zhang~\cite{ZhangKptDisc18} & 3.46 & 5.36\\
    Lorenz~\cite{Lorenz19} & 3.24 & 3.91 \\
    \midrule
    Ours & \textbf{2.66} & \textbf{3.77} \\
    \bottomrule\\[-1em]
    \end{tabular}}
    \end{center}\vspace{-0.7cm}
\end{table} 


\noindent\textbf{Restricted setting}
Table~\ref{table:kpt_restricted} shows the performance of keypoint estimation for the restricted setting. We use inter-ocular distance (IOD) error as a metric for MAFL and edge distance normalized error for CUB. Although our model uses the features learned from image-level supervision, our method shows the state-of-the-art performance on both datasets compared with the unsupervised methods. 



\noindent\textbf{Full dataset}
To show the sample efficiency of the representation from our model, we finetune the keypoint estimation network by varying the amount of keypoint annotation with 1\%, 10\%, and 100\%. We use the Percentage of Correct Keypoints (PCK) metric, which defines correct prediction if the distance between the ground truth and the prediction is within $\alpha=0.1$ with respect to the bounding box size. Figure~\ref{fig:diff_sup} shows the average PCK score over 3 different runs with supervised baseline GlobalNet~\cite{CPN17}, which is the same architecture for our keypoint module. Although the performance reaches almost the same after using 10\% of the data, our model shows better performances when there is an extremely limited amount of supervision. 





\begin{figure}
\centering
\includegraphics[width=8cm]{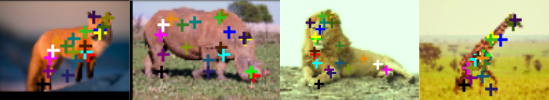}
\caption{Qualitative results for images with unseen categories}
\label{fig:unseen_main}\vspace{-0.2cm}
\end{figure}

\noindent\textbf{Keypoint discovery from unseen categories}
We show qualitative results on the animals from unseen categories in Figure~\ref{fig:unseen_main} using the model trained with the AnimalPose dataset. Since the species in AnimalPose contain diverse animals, our model can handle animals with various appearances across the species. We test on fox, rhino, lion, and giraffe images, which never appeared in the training dataset, and observe consistent part discovery across different species. 



\subsection{Downstream Tasks}

\begin{table}
    \caption{Performance on weak-supervision task (fine-grained classification) with our baseline ResNet-50~\cite{He2016DeepRL} with an image size of 128x128}
    \label{table:weak_supervision}
    \begin{center}
        \scalebox{0.9}{
        \begin{tabular}{llc}
        \toprule
            Method & Dataset & Accuracy \\
            \midrule
            ResNet-50~\cite{He2016DeepRL} (our baseline) & CUB & 67.9 \\
                Ours  & - & \textbf{68.9} \\
            ResNet-50~\cite{He2016DeepRL} (our baseline) & StanfordDogs & \textbf{71.5} \\
            Ours  & - & 69.7 \\
            \bottomrule\\[-1em]
        \end{tabular}
        }
    \end{center}\vspace{-0.8cm}
\end{table}

    

We show the performance of the fine-grained classification, which is the weak-supervision task, in Table~\ref{table:weak_supervision}. Since the goal of our method is to discover keypoints, this representation does not necessarily give the performance gain on all tasks. However, fine-grained classification on the CUB dataset shows improvement over the baseline, which is trained with the size of 128x128 images on ResNet-50~\cite{He2016DeepRL}.\\





\begin{figure}
    \centering
    \subfigure[keypoint annotation (TigDog)]{\includegraphics[width=4cm]{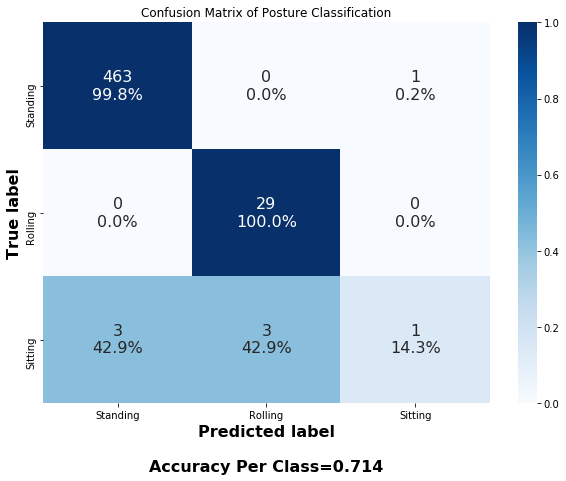}}
    \subfigure[Ours (TigDog)]{\includegraphics[width=4cm]{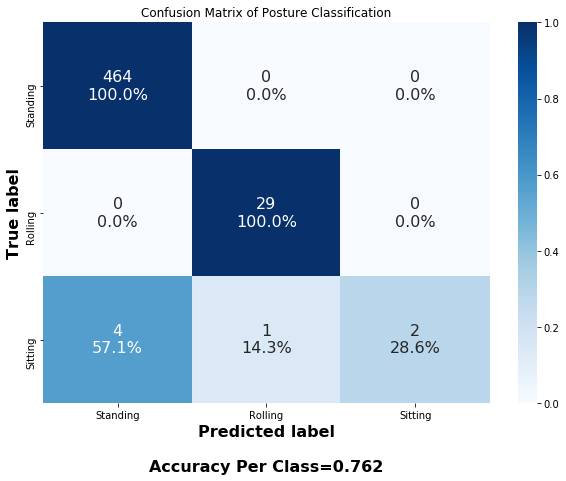}}
    \subfigure[Keypoint annotation (DogPart)]{\includegraphics[width=4cm]{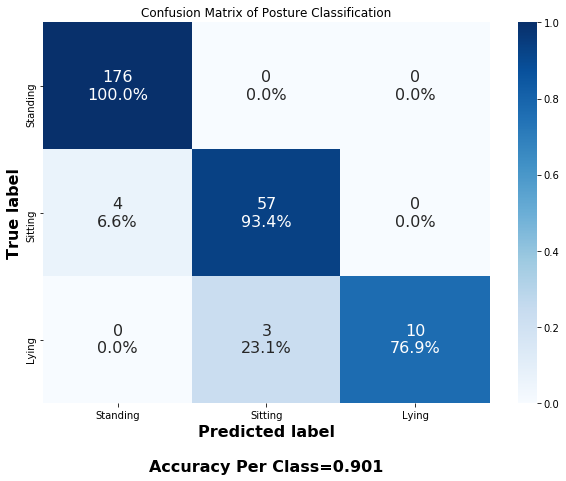}}
    \subfigure[Ours (DogPart)]{\includegraphics[width=4cm]{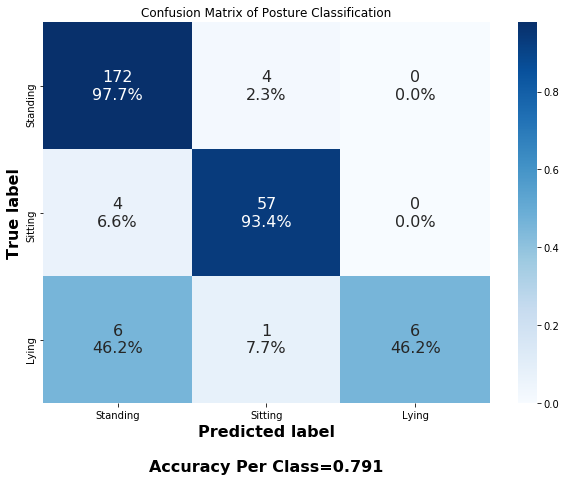}}
    \caption{Confusion matrix and per-class accuracy for posture-based action prediction of TigDog and DogPart datasets.}
    \label{fig:confusion_mat}\vspace{-0.2cm}
\end{figure}

\begin{figure}
\centering
\subfigure[Keypoint annotation]{\includegraphics[width=3.5cm]{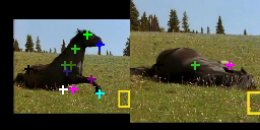}}
\subfigure[Ours]{\includegraphics[width=3.5cm]{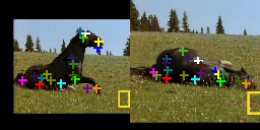}}
\caption{Ground truth annotation and discovered keypoints for sitting and lying actions from TigDog dataset.}
\label{fig:action_parts}\vspace{-0.2cm}
\end{figure}

\noindent\textbf{Posture-based activity prediction}
To demonstrate the representation power of the discovered keypoints, we directly apply the discovered keypoints, without any keypoint labels, to a simple behavior classification task (\eg sitting, standing, and lying) for two different datasets. We train two fully-connected layers with an input of the keypoint locations. For these experiments, we did not train the keypoint discovery model for each dataset due to the limited size of curated datasets. We used the model trained using AnimalPose~\cite{CaoICCV19} to TigDog~\cite{delpero15cvpr} and StanfordDogs~\cite{biggs2020wldo,Khosla_FGVC2011} to DogPart~\cite{Barnard2016} experiments, which further demonstrates the generalization ability of our trained models. Note that we do not use the regressed or finetuned keypoints. In Figure~\ref{fig:confusion_mat}, we provide the behavior classification results trained from human-annotated ground truth keypoints as a baseline, which is expected to be an upper bound performance. Our model achieves comparable performance in most of the categories. Surprisingly, our method on a simplified TigDog behavior task shows better performance. This is due to the scarce keypoint annotations for the occluded parts (Fig.~\ref{fig:action_parts}). Since our model always predicts all the parts around the animal location, it gives more information for the lying or rolling behaviors. 


\subsection{Discussion}

\begin{figure}
\centering
\includegraphics[width=8cm]{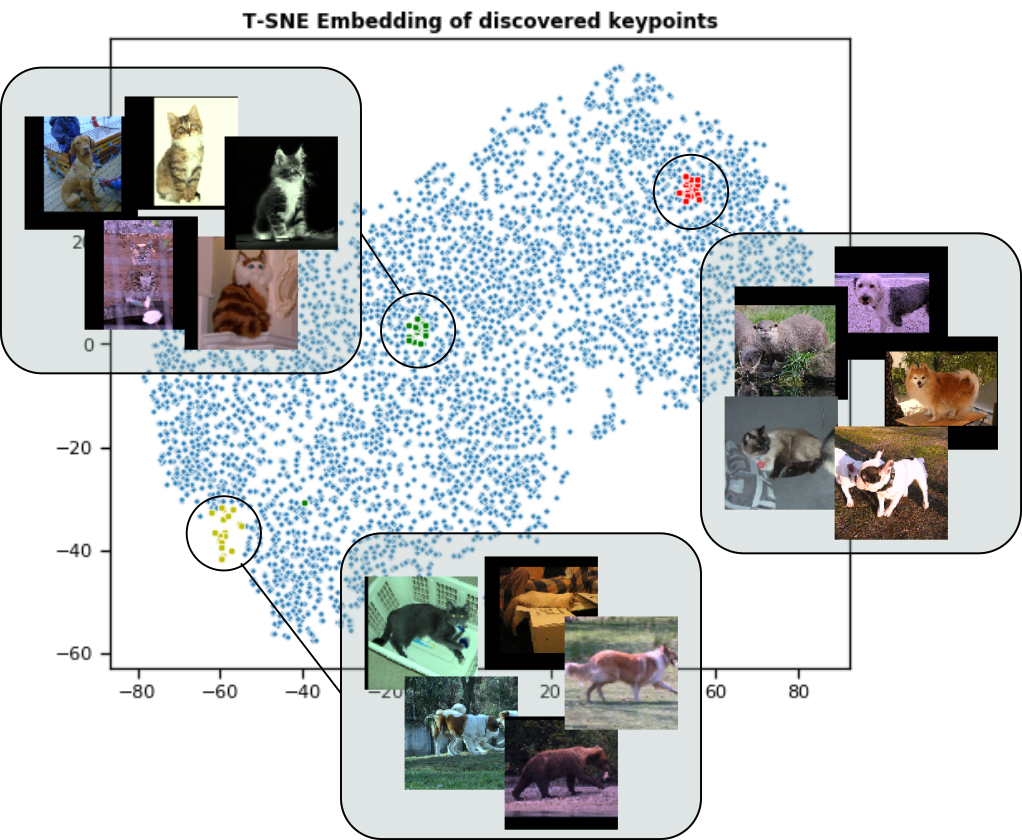}
\caption{Pose embedding for viewpoint-based equivariance}
\label{fig:pose_embedding}\vspace{-0.2cm}
\end{figure}

\noindent\textbf{Pose embedding}
Since our model uses the discovered keypoints to apply a viewpoint-based equivariance constraint, it is important to check whether the model can capture the viewpoint variation. We visualize the T-SNE~\cite{vanDerMaaten2008} embedding of the discovered keypoints from AnimalPose~\cite{CaoICCV19} dataset and the corresponding images from three different random locations. Embedding based on the discovered keypoints shows a high correlation with the viewpoint. 

\begin{figure}
\centering
\includegraphics[width=8cm]{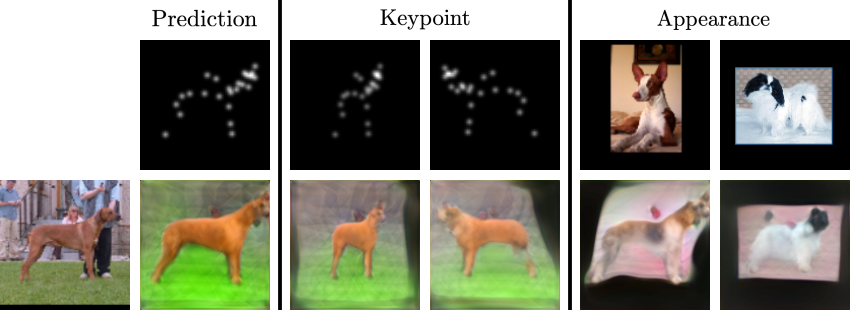}
\caption{Appearance and geometry factorization: We generate the image using the scaled and flipped keypoint as a geometry bottleneck with the input image appearance feature. Given the predicted keypoint from the input, we reconstruct the images using the appearance feature from the top right images.}
\label{fig:img_manipulation}\vspace{-0.2cm}
\end{figure}

\noindent\textbf{Appearance and geometry factorization}
Our model uses image reconstruction to discover the keypoints. Although image generation is not a primary goal of our method, our model can manipulate images with a huge viewpoint and appearance variation. We visualize the generated image given the geometry and the appearance bottleneck in Figure~\ref{fig:img_manipulation}. 

\begin{figure}
\centering
\includegraphics[width=8cm]{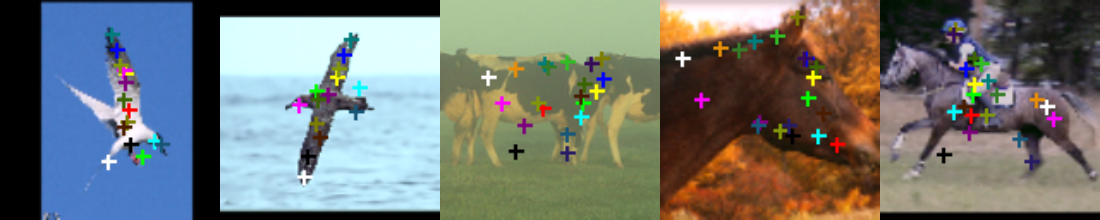}
\caption{Failure cases: Our model fails to discover consistent keypoints for the rare poses and cannot handle the missing or occluded parts. The spatial correlation problem still remains for the images with large appearance variations. }
\label{fig:failure}\vspace{-0.2cm}
\end{figure}

\noindent\textbf{Failure cases}
Figure~\ref{fig:failure} shows the failure cases. Since the discovered keypoints heavily depend on the training data distribution, our model struggles to predict the poses that do not frequently appear in the training data. Also, the model is trained to predict all the keypoint locations without considering the occlusion or missing parts. When the appearance variation is too high, the model fails to predict consistent keypoints.




\section{Conclusion}
We proposed a method to discover the keypoints from the images with various viewpoints by exploiting the weak labels. Our method can successfully discover keypoints when the target instances show large appearance and viewpoint variations. The proposed method has shown strong empirical results for the task of keypoint estimation with a limited amount of supervision. Furthermore, we demonstrated the representation power of our discovered keypoints by running an off-the-shelf model to downstream tasks from different datasets. 

{\small
\bibliographystyle{ieee_fullname}
\bibliography{egbib}
}

\newpage
\setcounter{section}{0}
\setcounter{equation}{0}
\setcounter{figure}{0}
\setcounter{table}{0}
\renewcommand{\theequation}{A-\arabic{equation}}
\renewcommand{\thefigure}{A-\arabic{figure}}
\renewcommand{\thetable}{A-\arabic{figure}}
\renewcommand\thesubsection{A-\arabic{subsection}}

\section*{Appendix}
\subsection{Qualitative results}
\begin{figure}[h]
    \centering
    \includegraphics[width=8cm]{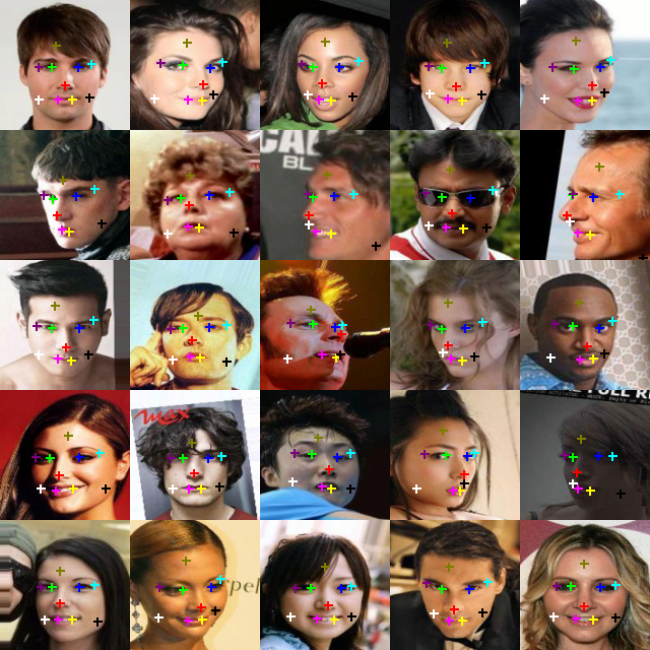}
    \caption{Qualitative results on CelebA with 10 discovered keypoints.}
    \label{fig:celeba}
\end{figure}

\begin{figure}[h]
    \centering
    \includegraphics[width=8cm]{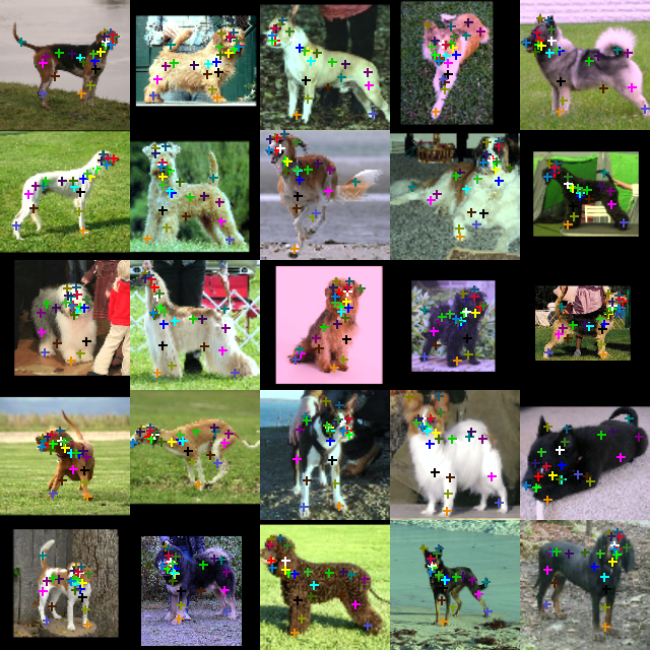}
    \caption{Qualitative results on StanfordDogs with 24 discovered keypoints.}
    \label{fig:stanforddogs}
\end{figure}

\begin{figure}[h]
    \centering
    \includegraphics[width=8cm]{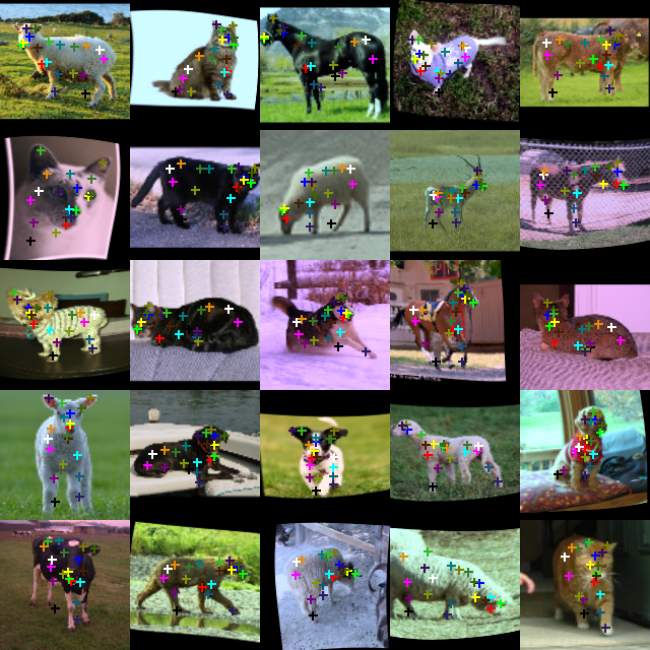}
    \caption{Qualitative results on AnimalPose with 20 discovered keypoints.}
    \label{fig:animalpose}
\end{figure}

\begin{figure}[h]
    \centering
    \includegraphics[width=8cm]{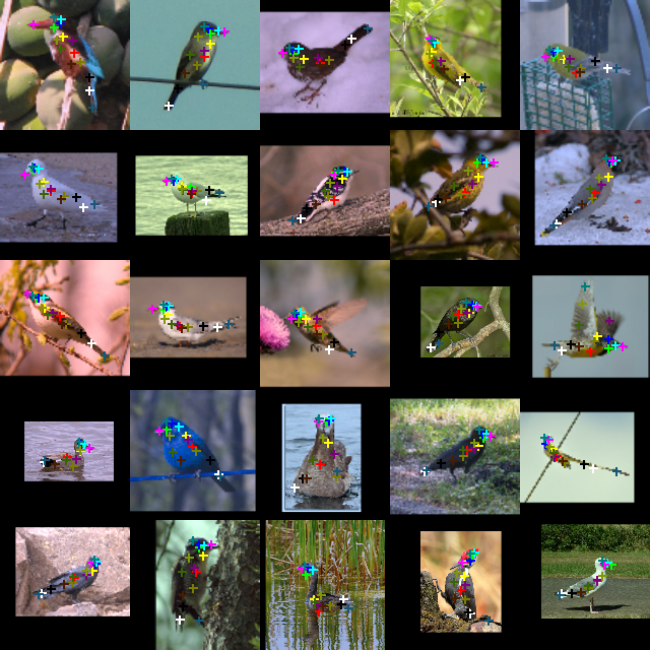}
    \caption{Qualitative results on CUB with 15 discovered keypoints.}
    \label{fig:cub}
\end{figure}

\begin{figure}[h]
    \centering
    \includegraphics[width=8cm]{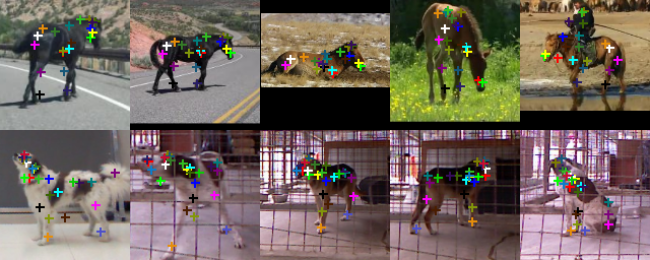}
    \caption{Qualitative results on TigDog and DogPart from the model trained on AnimalPose and StanfordDogs, respectively.}
    \label{fig:tigdog_dogpart}
\end{figure}

\begin{figure}[h]
    \centering
    \includegraphics[width=8cm]{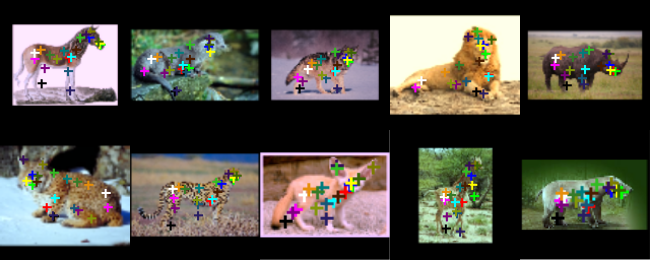}
    \caption{Qualitative results on unseen categories from the model trained on AnimalPose.}
    \label{fig:unseen}
\end{figure}

We provide additional qualitative results on the following datasets: CelebA (Fig~\ref{fig:celeba}), StanfordDogs (Fig~\ref{fig:stanforddogs}), AnimalPose (Fig~\ref{fig:animalpose}), and CUB (Fig~\ref{fig:cub}). To show the robustness of our trained model, we visualize the discovered keypoints on DogPart and TigDog datasets (Fig~\ref{fig:tigdog_dogpart}), where the model was originally trained on StanfordDogs and AnimalPose datasets, respectively. We also test our method on the images with unseen categories in Fig~\ref{fig:unseen}. Our method can reliably discover keypoints when the instances have large shape and viewpoint variations. However, since our model is trained to predict all the keypoints without considering the presence of occlusion or missing parts, the discovered keypoints from the cropped images are partially mapped to look-alike parts. Also, the model cannot distinguish the front and the back legs for the images with front-facing animals. 

\begin{figure}[h]
    \centering
    \includegraphics[width=8cm]{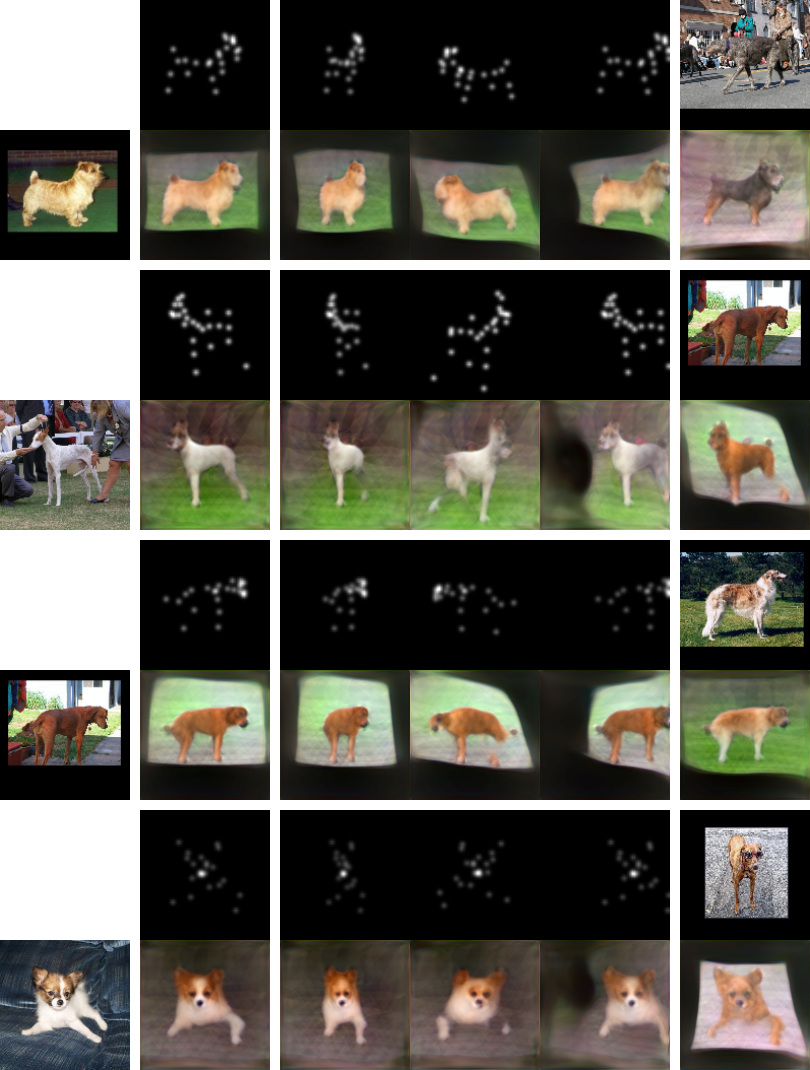}
    \caption{Appearance and geometry factorization on StanfordDogs dataset.}
    \label{fig:manipulation}
\end{figure}

\begin{figure}[h]
    \centering
    \includegraphics[width=8cm]{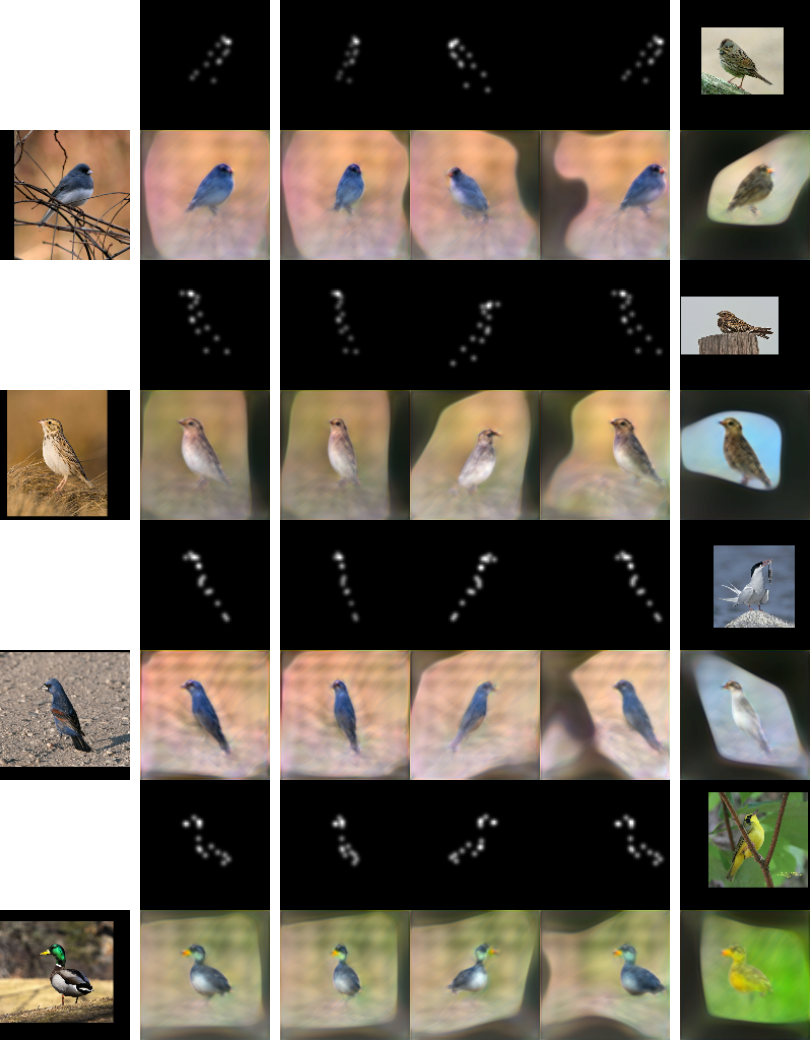}
    \caption{Appearance and geometry factorization on CUB dataset.}
    \label{fig:manipulation_cub}
\end{figure}

\subsection{Appearance and geometry factorization}
Our model uses image generation to discover the keypoints by disentangling the appearance and geometry features. Though image reconstruction is not a primary goal of our work, our method can generate images with large appearance and viewpoint variations. We visualize the generated images by scaling, flipping, and moving the geometry bottleneck on StanfordDogs (Fig~\ref{fig:manipulation}) and CUB (Fig~\ref{fig:manipulation_cub}) datasets. We also show the generated images using the appearance feature from the top images with the same geometry bottleneck.

\subsection{Implementation details}
Here we provide the architecture details about the keypoint, reconstruction, and weak supervision modules. We specify the block type and the feature dimension of each layer in Tables~\ref{table:kpt_module} to~\ref{table:weak_module}. 

\subsection{Keypoint module}

\begin{figure}[h]
\centering
    \subfigure[Lateral]{\includegraphics[width=2cm]{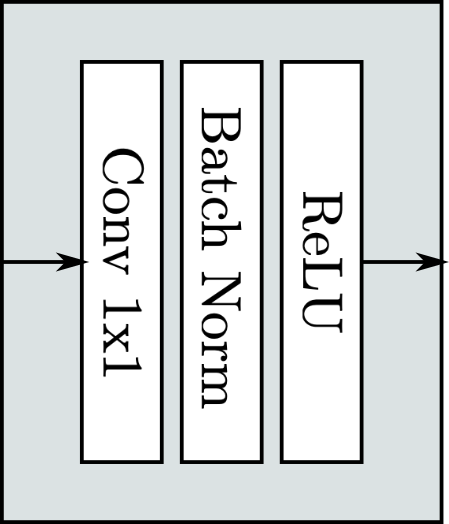}}\hspace{1cm}
    \subfigure[Upsample]{\includegraphics[width=2cm]{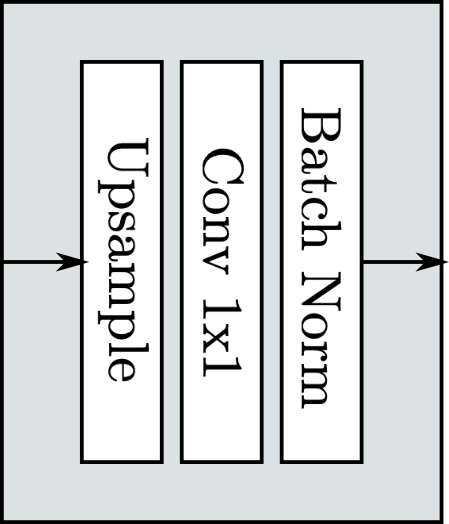}}
    \subfigure[Predict]{\includegraphics[width=4cm]{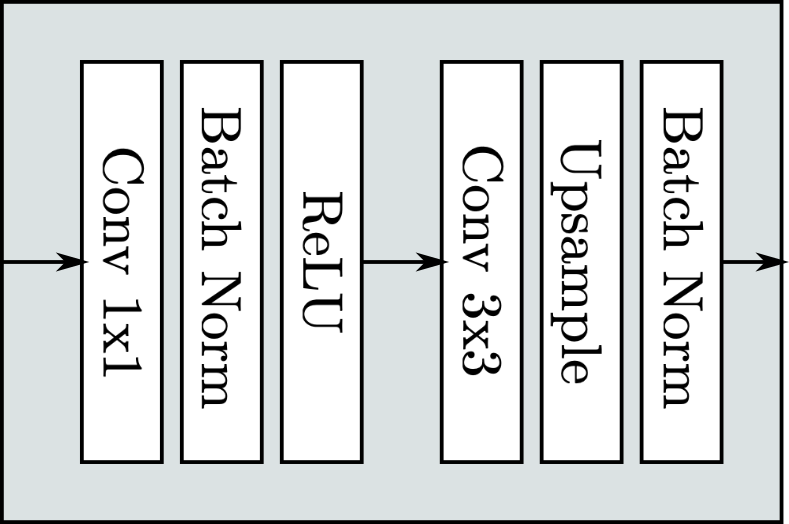}}
    \caption{Layer description on lateral, upsample, and predict blocks.}
    \label{fig:lateral_up}
\end{figure}

\begin{table}[h]
    \caption{Architecture details about the keypoint module.}
    \label{table:kpt_module}
    \begin{center}
        \scalebox{0.8}{
        \begin{tabular}{lcccc}
        \toprule
            Type & input\_dim & out\_ch\_dim & output\_size  \\
            \midrule
            Lateral & 2048 & 256 & 4x4 \\
            Upsample & 256 & 256 & 8x8 \\
            Predict & 256 & \# parts & 64x64 \\
            Lateral & 1024 & 256 & 8x8 \\
            Upsample & 256 & 256 & 16x16 \\
            Predict & 256 & \# parts & 64x64 \\
            Lateral & 512 & 256 & 16x16 \\
            Upsample & 256 & 256 & 32x32 \\
            Predict & 256 & \# parts & 64x64 \\
            Lateral & 256 & 256 & 32x32 \\
            Upsample & 256 & 256 & 64x64 \\
            Predict & 256 & \# parts & 64x64 \\
            \bottomrule\\[-1em]
        \end{tabular}
        }
    \end{center}
\end{table}

Figure~\ref{fig:lateral_up} shows the description of the lateral, upsample, and predict blocks that are used for the keypoint module. For the convolution, the stride is always set to 1. Note that the input for each lateral block is the output from the convolution block of the ResNet-50 encoder. The output from the previous lateral block is added to the next lateral block so that the network does not lose the semantic information from the deep layers and spatial information from the shallow layers. The final heatmap, which is the output from the final predict block, is used to generate the geometry bottleneck. 

\subsection{Reconstruction module}

\begin{figure}[h]
    \centering
    \includegraphics[width=4cm]{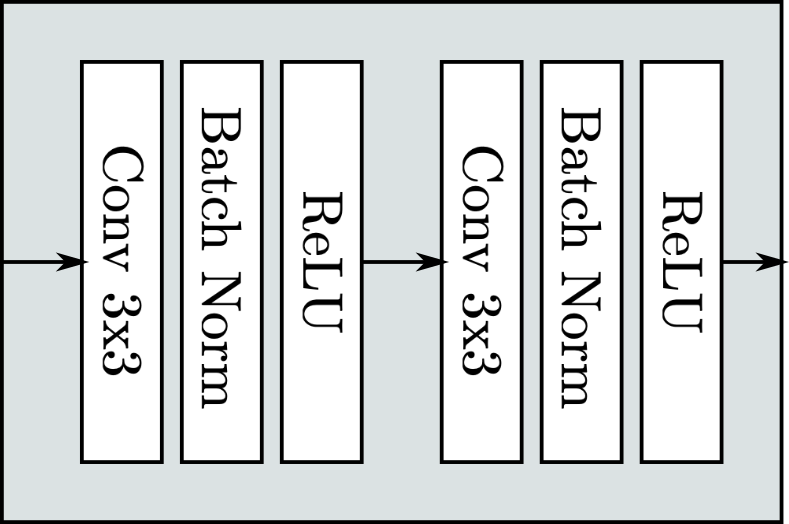}
    \caption{Basic convolution block which is used for reconstruction module.}
    \label{fig:lateral}
\end{figure}

\begin{figure}[h]
    \centering
    \includegraphics[width=8cm]{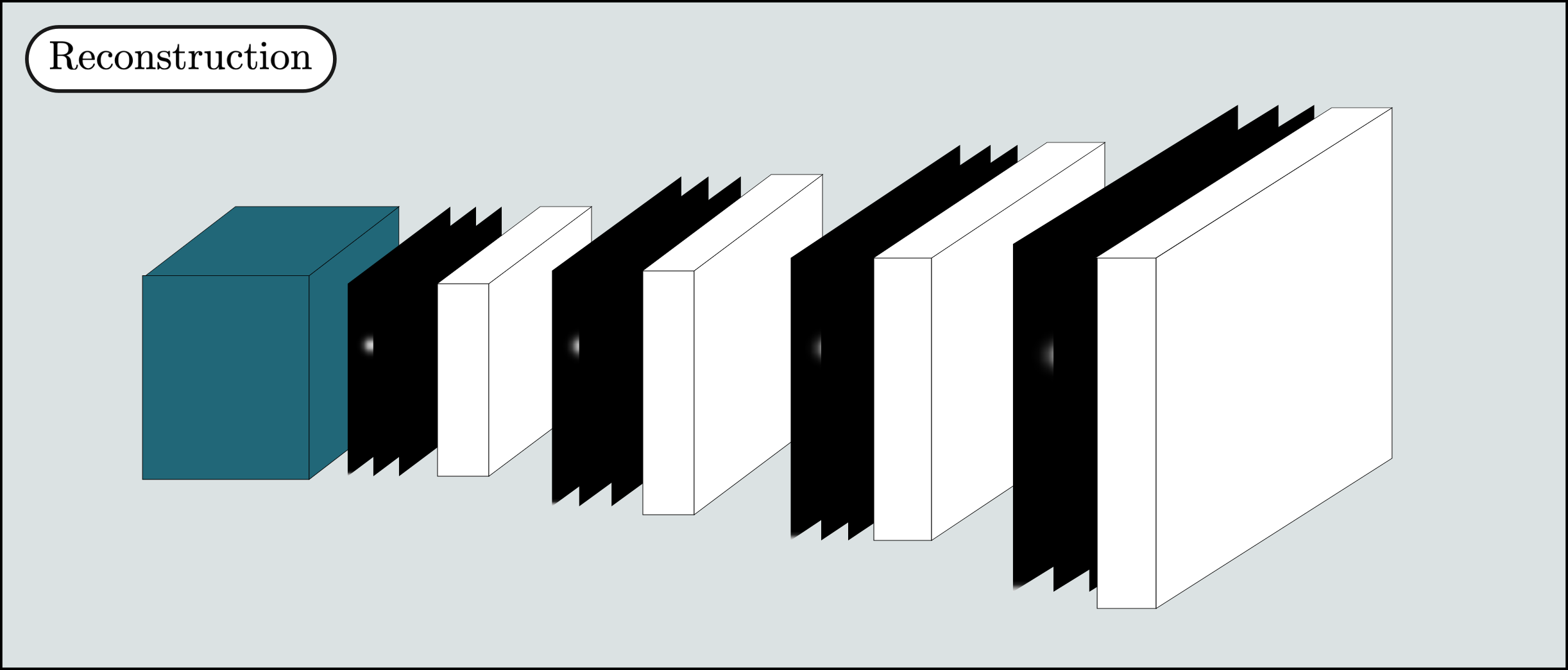}
    \caption{Reconstruction module.}
    \label{fig:recon}
\end{figure}

We feed the geometry bottleneck to each resolution in the reconstruction module. In this procedure, we use a different normalization factor for generating the gaussian heatmap bottleneck so that the layers from higher resolution get more concentrated geometry features. We use $[0.1, 0.1, 0.01, 0.01, 0.001]$ to the $\sigma$ variable in Eq 1 for each resolution. Figure~\ref{fig:lateral} shows the basic convolution block (conv\_block in Table~\ref{table:recon_module}), which is composed of 3x3 convolution, batch normalization, and ReLU activation. The upsampling layer in Table~\ref{table:recon_module} is a single upsampling layer which is different from the upsampling block in Table~\ref{table:kpt_module}.

\begin{table}[h]
    \caption{Architecture details about the reconstruction module.}
    \label{table:recon_module}
    \begin{center}
        \scalebox{0.8}{
        \begin{tabular}{lcccc}
        \toprule
            Type & input\_ch\_dim & kernel\_size & out\_ch\_dim & output\_size  \\
            \midrule
            Upsampling & - & - & - & 8x8 \\
            Conv\_block & 2048+\# parts & - & 1024 & 8x8 \\
            Upsampling & - & - & - & 16x16 \\
            Conv\_block & 1024+\# parts & - & 512 & 16x16  \\
            Upsampling & - & - & - & 32x32 \\
            Conv\_block & 512+\# parts & - & 256 & 32x32  \\
            Upsampling & - & - & - & 64x64 \\
            Conv\_block & 256+\# parts & - & 128 & 64x64  \\
            Upsampling & - & - & - & 128x128 \\
            Conv\_block & 128+\# parts & - & 64 & 128x128 \\
            Convolution & 64 & 1 & 3 & 128x128 \\
            \bottomrule\\[-1em]
        \end{tabular}
        }
    \end{center}
\end{table}

\subsection{Weak supervision module}

\begin{figure}[h]
    \centering
    \includegraphics[width=4cm]{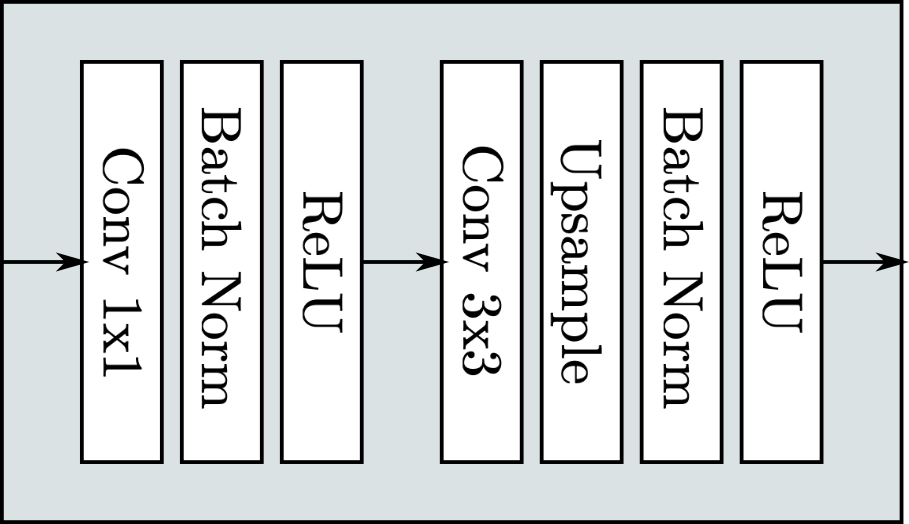}
    \caption{Layer specification in the convolution block (conv\_block\_w) in weak supervision module.}
    \label{fig:conv_block_weak}
\end{figure}

Figure~\ref{fig:conv_block_weak} shows the layer description of the convolution block in the weak supervision module. The input for each convolution block is the output from the ResNet-50 encoder. We generate the base feature for the final classification task using the concatenation of the features from the module in Table~\ref{table:weak_module}.

\begin{table}[h]
    \caption{Architecture details about the weak supervision module.}
    \label{table:weak_module}
    \begin{center}
        \scalebox{0.8}{
        \begin{tabular}{lccc}
        \toprule
            Type & input\_ch\_dim & out\_ch\_dim & output\_size  \\
            \midrule
            Conv\_block\_w & 2048 & 256 & 64x64 \\
            Conv\_block\_w & 1024 & 256 & 64x64 \\
            Conv\_block\_w & 512 & 256 & 64x64 \\
            Conv\_block\_w & 256 & 256 & 64x64 \\
            \bottomrule\\[-1em]
        \end{tabular}
        }
    \end{center}
\end{table}

\subsection{Dataset}
\noindent\textbf{TigDog} We use the subset of the TigDog dataset for the posture-based action classification task, where the action category consists of sitting, rolling, and standing. Most of the activity classes require temporal information, thus we used the images from the category of \textit{walking}, \textit{rolling}, \textit{standing up} and \textit{sitting}, where the action can be identified by watching a single frame. From the frames of \textit{standing up} and \textit{sitting} behaviors, we manually selected the images that have sitting posture for the sitting category. \\

\subsection{Hyperparameter settings}
Our method does not require an extensive hyperparameter search. The experimental results were obtained by applying the same weight for each loss. We show the starting epoch number for the curriculum learning of the viewpoint equivariance loss in Table~\ref{table:hyperparameter}. We use the SGD optimizer for all the experiments.  

\begin{table}[h]
    \caption{Hyperparameter settings for each dataset.}
    \label{table:hyperparameter}
    \begin{center}
        \scalebox{0.9}{
        \begin{tabular}{lcccccc}
        \toprule
            Dataset & lr & $L_{perc} $ & $L_w$ & $L_v$ & epoch & \# parts \\
            \midrule
            CelebA & 0.001 & 1 & - & - & - & 10\\
            CUB & 0.001 & 1 & 1 & 1 & 30 & 15 \\
            AnimalPose & 0.001 & 1 & 1 & 1 & 40 & 20 \\
            StanfordDogs & 0.001 & 1 & 1 & 1 & 30 & 24 \\
            \bottomrule\\[-1em]
        \end{tabular}
        }
    \end{center}\vspace{-0.8cm}
\end{table}
\end{document}